\pdfoutput=1

\documentclass[11pt]{article}

\usepackage[]{ACL2023}

\usepackage{times}
\usepackage{latexsym}

\usepackage[T1]{fontenc}

\usepackage[utf8]{inputenc}

\usepackage{microtype}

\usepackage{inconsolata}
\usepackage{booktabs}
\usepackage{multirow}
\usepackage{multicol}
\usepackage{url}
\usepackage{hyperref}
\usepackage{graphicx}
\usepackage{algorithm}
\usepackage{algorithmic}
\usepackage{amsmath}
\usepackage{amssymb}
\usepackage{bm}
\title{Domain Incremental Lifelong Learning in an Open World}

\author{
Yi Dai$^{1}$\thanks{\quad Work done while the author was interning at Alibaba.} \thanks{\quad Equal contribution.}, \
Hao Lang$^{2}$\footnotemark[2], \
Yinhe Zheng$^{2}$\thanks{\quad \small Corresponding author.}, \
Bowen Yu$^{2}$, \
Fei Huang$^{2}$, \
Yongbin Li$^{2}$\footnotemark[3] \\
$^1$ Department of Computer Science and Technology, Tsinghua University
$^{2}$ Alibaba Group\\
\small \texttt{\{hao.lang, yubowen.ybw, f.huang, shuide.lyb\}@alibaba-inc.com}, \\
\small \quad \texttt{dai-y21@mails.tsinghua.edu.cn, zhengyinhe1@163.com}\\
}

\begin{document}
\maketitle
\begin{abstract}
Lifelong learning (LL) is an important ability for NLP models to learn new tasks continuously. Architecture-based approaches are reported to be effective implementations for LL models. However, it is non-trivial to extend previous approaches to domain incremental LL scenarios since they either require access to task identities in the testing phase or cannot handle samples from unseen tasks. In this paper, we propose \textbf{Diana}: a \underline{d}ynam\underline{i}c \underline{a}rchitecture-based lifelo\underline{n}g le\underline{a}rning model that tries to learn a sequence of tasks with a prompt-enhanced language model. Four types of hierarchically organized prompts are used in Diana to capture knowledge from different granularities. Specifically, we dedicate task-level prompts to capture task-specific knowledge to retain high LL performances and maintain instance-level prompts to learn knowledge shared across input samples to improve the model's generalization performance. Moreover, we dedicate separate prompts to explicitly model unseen tasks and introduce a set of prompt key vectors to facilitate knowledge sharing between tasks. Extensive experiments demonstrate that Diana outperforms state-of-the-art LL models, especially in handling unseen tasks. We release the code and data at \url{https://github.com/AlibabaResearch/DAMO-ConvAI/tree/main/diana}.
\end{abstract}

\section{Introduction}

An essential ability of humans is to learn new tasks continuously in their lifetime since our surrounding world is ever involving \cite{thrun1995lifelong}.
Humans need to learn inputs from unseen new tasks everyday.
However, neural network based NLP models tend to rapidly lose previously acquired knowledge when trained on new tasks.
This phenomenon is referred to as catastrophic forgetting \cite{french1999catastrophic},
and it’s important to equip NLP models with the lifelong learning (LL) ability to alleviate this issue in advanced AI applications.


An effective method to build LL models is the architecture-based approach~\cite{chen2016net2net,rusu2016progressive,fernando2017pathnet,chayut2020},
in which task-specific components are used to isolate knowledge for each separate task \cite{mancini2018adding}. Recently, to leverage the power of pre-trained language model (PLM), some architecture-based LL models convert NLP tasks into a unified language modeling (LM) format \cite{sanh2021multitask,xie2022unifiedskg} and learn these tasks using a PLM. Separate prompts \cite{qin2022lfpt} or adapters \cite{madotto2021continual} are allocated for different tasks to avoid the catastrophic forgetting issue.\looseness=-1


\begin{figure}[t]
\centering
\includegraphics[width = 1.0\linewidth]{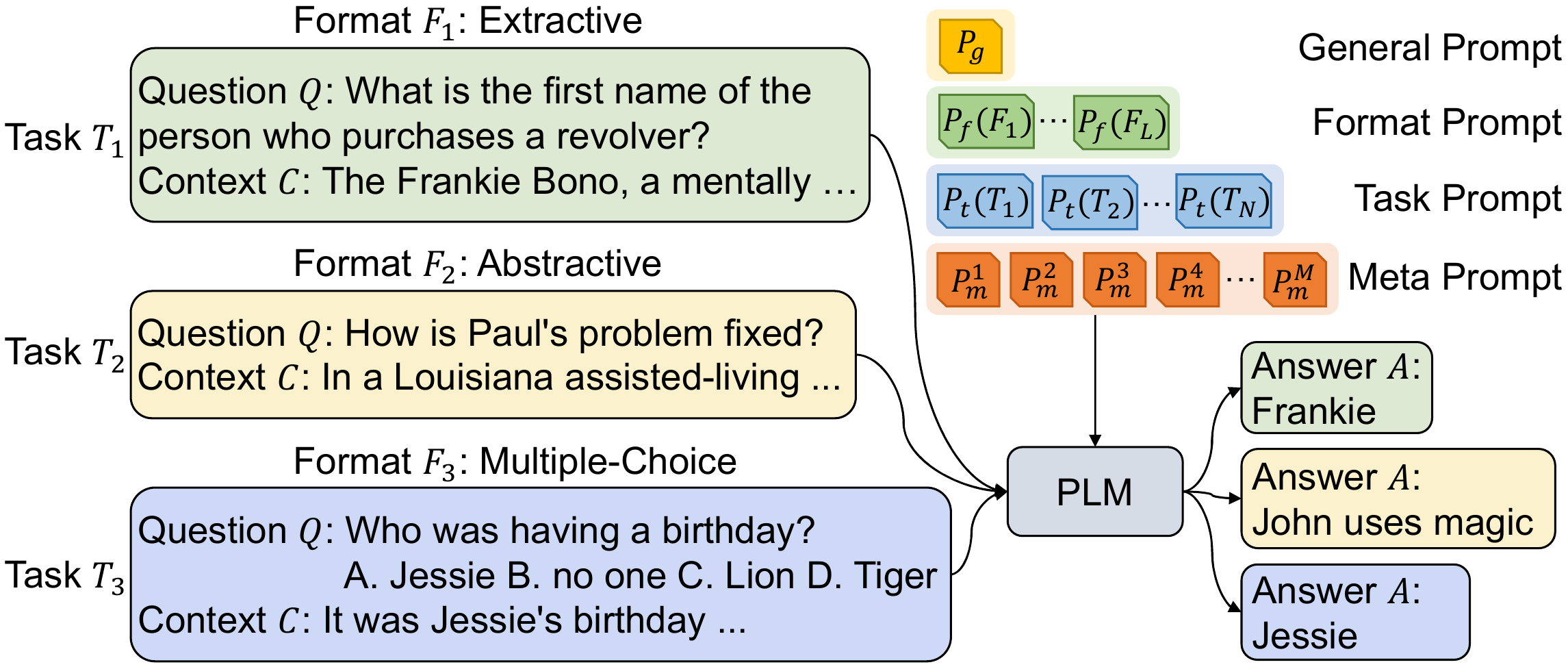}
\caption{An overview of Diana. A pre-trained language model is used to learn tasks in different formats with hierarchically organized prompts.}
\label{fig:framework}
\end{figure}

However, despite the reported effectiveness, most above models are designed for the task incremental learning scenario,
in which we assume task IDs for testing samples are available~\cite{wang2022dualprompt,Wang_2022_CVPR}.
This setting limits the application of LL models
because practical applications usually follow a more general domain incremental learning scenario \cite{van2022three},
i.e., we cannot access the task IDs of most input samples.

There are generally two approaches to building LL models for domain incremental learning.
One is to predict the task ID of each testing sample~\cite{wortsman2020sup},
and activate specified components based on the prediction (Figure \ref{fig:unified}a). This scheme achieves high LL performances if the predicted ID is correct \cite{madotto-etal-2021-continual}. 
However, these models cannot handle samples from unseen tasks since there are no components designated for these samples and thus no task IDs to be predicted.
This hinders the application of LL models because we often encounter samples from unseen tasks in practical situations \cite{dietterich2017steps}.


Another approach to building domain incremental LL models is to organize model components at the instance-level, i.e., a pool of fine-grained components are dynamically combined in the forward pass for each input instance (Figure \ref{fig:unified}b).
This approach avoids the trouble of explicitly determining task IDs.
However, it usually yields low LL performance because there are no dedicated components for each task to capture task-specific knowledge~\cite{wang2022dualprompt}.  


In this study, we combine the advantages of the above two approaches and propose \textbf{Diana}: a \underline{d}ynam\underline{i}c \underline{a}rchitecture-based lifelo\underline{n}g le\underline{a}rning model. We convert different NLP tasks into a unified LM format and propose to learn these tasks using a prompt-enhanced PLM (Figure \ref{fig:framework}). Specifically, Diana maintains four types of prompts to capture task knowledge from different granularities: 1. A \emph{general prompt} $P_g$ is used for all tasks; 2. The \emph{format prompt}s $P_f$ are shared between tasks in a similar format; 3. A \emph{task prompt} $P_t$ is assigned for each incoming task; 4. A pool of \emph{meta prompt}s $P_m$ are dynamically combined for each input instance. These four types of prompts present a hierarchical structure with a decreasing knowledge granularity, i.e., $P_g$ captures global knowledge between all tasks, while $P_m$ captures local knowledge that is shared between instances.

Diana can better generalize to unseen tasks while achieving high LL performances since its components are organized at both task and instance level. Moreover, we also maintain key vectors for $P_t$ and $P_m$ to better share task knowledge, and allocate separate task prompts to explicitly model samples for unseen tasks.
Extensive experiments on benchmark NLP tasks indicate that Diana outperforms state-of-the-art (SOTA) baselines, especially in handling unseen tasks. Our main contributions are:

1. We propose Diana: a novel architecture-based domain incremental LL model that uses hierarchically organized prompts to capture knowledge in different granularities. 

2. We are the first to consider unseen tasks in the testing phase of LL models. Specific prompts are designated in Diana to handle unseen tasks, and prompt keys are built to facilitate sharing of task knowledge.\looseness=-1
  
3. Extensive experiments show that Diana outperformed SOTA baselines.

\section{Related Work}
\textbf{Lifelong Learning} aims at incrementally acquiring new knowledge without catastrophically forgetting previously learned ones. Generally, three categories of LL methods are proposed: \textbf{1.} Rehearsal-based methods~\cite{rebuffi2017icarl,shin2017continual,sun2019lamol,chaudhry2019efficient,buzzega2020dark} preserve past knowledge by replaying data from learned tasks; \textbf{2.} Regularization-based methods~\cite{kirkpatrick2017overcoming,zenke2017continual,li2017learning,ritter2018online,pmlr-v108-farajtabar20a} consolidate model parameters that are important to previous tasks by introducing additional regularization terms; \textbf{3.} Architecture-based methods~\cite{chen2016net2net,rusu2016progressive,fernando2017pathnet,maltoni2019continuous} add task-specific parameters to an existing base model for each task to prevent forgetting.

Experiment settings of LL methods can be generally classified into three scenarios based on whether the task ID is provided for testing samples and whether it must be inferred~\cite{van2019three}, i.e., task-incremental learning~\cite{mallya2018packnet,ebrahimi2020adversarial}, domain-incremental learning~\cite{pu2021lifelong,gao2022forget}, and class-incremental learning~\cite{zhang2020class}. In this work, we focus on the domain-incremental learning setting, where task ID is not provided for each testing sample. One line of methods in this category attempt to detect the task ID for each input sample~\cite{madotto-etal-2021-continual}. However, these methods fail to generalize to unseen tasks \cite{wang2022dualprompt}. Another line of methods try to build a dynamic architecture for each input sample, for example, maintaining a pool of prompts that can be dynamically combined \cite{Wang_2022_CVPR}. However, these methods yield sub-optimal performance since no task-specific parameters are used. Our model Diana is the first attempt to take advantage of the two aforementioned types of methods.\looseness=-1


\textbf{Pre-trained LM}
is becoming the de facto standard component for NLP models.
To encourage knowledge sharing, existing approaches attempt to cast all NLP tasks into a unified text-to-text format~\cite{mccann2019the} and learn these tasks by finetuning a PLM. A similar work compared to ours is ProQA \cite{zhong2022proqa}, in which different QA tasks are unified and a set of structured prompts are used. However, ProQA only considers two QA tasks and is limited to the task incremental learning scenario, while our model is designed to tackle more general NLP tasks in a more general domain incremental learning scenario.

\section{Method}
\label{sec:method}
\subsection{Task Formulation}
In this study, we aim to sequentially learn $N$ tasks $T_1, \cdots , T_N$ that are presented in $L$ different formats $F_1, \cdots, F_L$, $(L \leq N)$. Each task $T_i$ is presented in a specific format $F_j$ (such as ``Classification'' or ``Summarization''), and each training sample of $T_i$ is a tuple of a context $C$, a question $Q$, and an answer $A$: ($C, Q, A$). Note that the format of each task can be easily inferred from the context-question pair ($C, Q$). Our model $g_\theta$ is built to predict $A$ based on $C$ and $Q$. We also consider a more challenging open domain lifelong learning setting, i.e., the model needs to predict answers for unseen tasks. Therefore, we collect another $N'$ unseen tasks $T_{N+1}, \cdots, T_{N+N'}$ that are only used for testing. We assume that all task identities of inputs are not available in the testing phase.\looseness=-1

\begin{figure}[t]
\centering
\includegraphics[width = 1.0\linewidth]{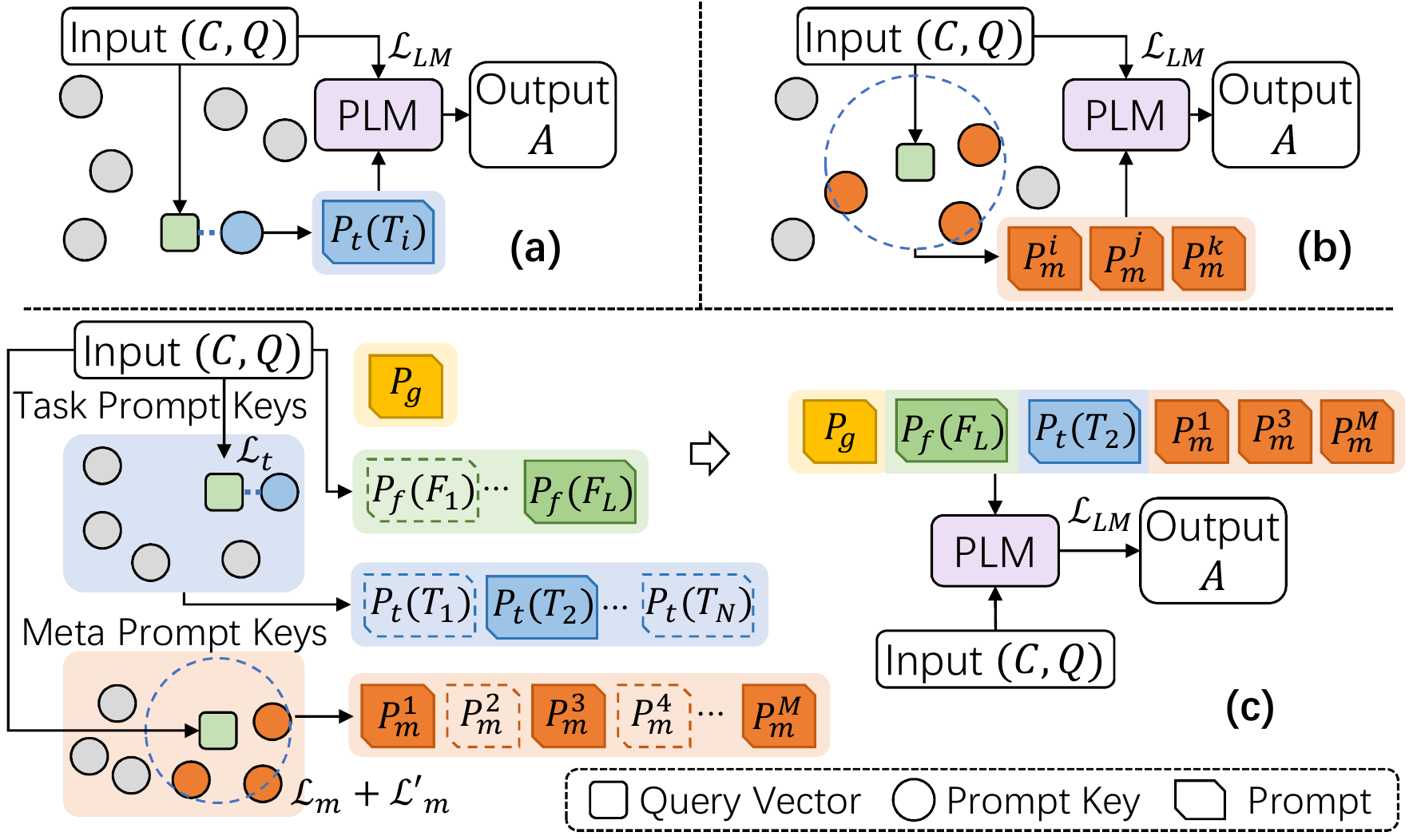}
\caption{Different prompt organization schemes.
(a) Each task is assigned a separate prompt and the closest prompt to the query vector is activated.
(b) A pool of prompts are maintained and the top-$M'$ closest prompts to the query vector are activated and combined.
(c) Four kinds of prompts are hierarchically organized and combined based on the task format and distances between the query vector and prompt keys.}
\label{fig:unified}
\end{figure}

\subsection{Framework of Hierarchical Prompts}
We follow previous approaches to serialize the context $C$, question $Q$, and answer $A$ into text sequences~\cite{khashabi-etal-2020-unifiedqa,zhong2022proqa} and use a prompt-enhanced encoder-decoder model $g_{\theta}$ to learn each task $T_i$ in Diana. We use soft prompts~\cite{liu2021gpt,lester-etal-2021-power,vu-etal-2022-spot} in our study, i.e., each prompt is a sequence of trainable embeddings that are randomly initialized and learned in the training process. For each training sample $(C, Q, A)$ from task ${T}_i$, we first construct a prompt $P(C, Q)$ based on ($C, Q$). Then the encoder takes in the concatenation of $P(C, Q)$, $C$, and $Q$ and the decoder predicts $A$, i.e., $A = g_{\theta}([P(C, Q);C;Q])$, in which ``$[;]$'' denotes the sequence concatenation operation.

Four types of prompts are contained in $P(C, Q)$, i.e., $P(C, Q) = [P_g;P_f(F_j);P_t(T_i);P_m(C, Q)]$ (Figure \ref{fig:unified}c). Specifically, $P_{g}$ is a general prompt, $P_f(F_j)$ is a format prompt (where $F_j$ is the format of task $T_i$), $P_t(T_i)$ is a task prompt and $P_m(C, Q)$ is a combined meta prompt. These four types of prompts are organized hierarchically so that they are shared by samples in different granularities:

\textbf{1. General Prompt} $P_{g}$ is shared for all training tasks so that it encodes global task knowledge.

\textbf{2. Format Prompt} $P_f(F_j)$ is shared between tasks in the same format $F_j$ so that it captures format-related knowledge, i.e., knowledge that is shared between tasks in the format $F_j$.

\textbf{3. Task Prompt} $P_t(T_i)$ is specifically allocated for the task $T_i$ and it is only shared for samples from $T_i$.
We use $P_t(T_i)$ to learn task-specific knowledge.
Moreover, to explicitly model samples from unseen tasks, we enlarge the set of task prompts with $L$ extra prompts $\hat{P_t}(F_1)$, $\cdots$, $\hat{P_t}(F_L)$, in which each prompt $\hat{P_t}(F_j)$ models the unseen task for a particular format $F_j$.

\textbf{4.~Meta Prompt} $P_m(C, Q)$ is a dynamic combination of various instance-level prompts.
Specifically, we maintain $M$ instance-level meta prompts $\{P_m^i\}_{i=1}^M$ and dynamically combine these prompts based on the ($C, Q$) to obtain $P_m(C, Q)$. 
$P_m(C, Q)$ captures the knowledge shared between similar training instances.


We expect these four types of prompts can capture knowledge from different granularities since they are shared in different scopes. 
Moreover, to facilitate knowledge sharing, we allocate a key vector $\bm{k}_t(T_i)$ and $\bm{k}_m^j$ to each task prompt $P_t(T_i)$ and meta prompt $P_m^j$, respectively, and build a fixed text encoder $h$ to map a context-question pair ($C, Q$) to a query vector $\bm{q}=h(C, Q)$. A two-stage learning process is introduced in Diana to learn these keys and $P(C, Q)$. Specifically, the first stage focuses on learning a representation space for prompt keys so that we can determine proper prompts to construct $P(C, Q)$. The second stage optimizes the constructed prompt $P(C, Q)$ and the backbone language model. These two stages are detailed in the following sections.



\subsection{Key Vector Space Learning}
\label{sec:keyspace}
We first optimize key vectors assigned to each task prompt and meta prompt to construct the prompt $P(C, Q)$ for each input ($C, Q$). Note that these key vectors are only used to determine the task prompt and meta prompt in $P(C, Q)$ because the general prompt $P_g$ is shared by all tasks in Diana, and the format prompt $P_f(F_j)$ can be determined based on the format of $C$ and $Q$ directly.

\textbf{Task Prompt Keys} 
help to determine the task prompt in $P(C, Q)$. Specifically, for a given input ($C, Q$), we first calculate its query vector $\bm{q}$ and then determine the most similar task prompt key $\bm{k}_t(T_i)$ to $\bm{q}$. The task prompt $P_t(T_i)$ associated with $\bm{k}_t(T_i)$ is used to construct $P(C, Q)$. 

Ideally, the key vector $\bm{k}_t(T_i)$ for a task prompt $P_t(T_i)$ should be located near samples from task $T_i$ and distant to samples from other tasks $T_j$ ($j\neq i)$. Therefore, when learning each task $T_i$, we maintain a small memory buffer $\mathcal{M}$ for samples from previously learned tasks $T_j$, ($j<i$), and design the following exponential angular triplet loss \cite{biexp-ye} to enforce the above property:
\begin{equation}
\small
\begin{aligned}
    \mathcal{L}_{t} = &{\rm exp}(||h(C,Q),\bm{k}_t(T_i)|| + \\
    &{\rm max}(1-||h(C_n, Q_n), \bm{k}_t(T_i)||,0)), 
\end{aligned}
    \label{eq:metric_learning}
\end{equation}
in which the operator $||\cdot,\cdot||$ determines the distance between two input vectors (here we use cosine distance), ($C_n$, $Q_n$) is a negative sample extracted from the memory buffer $\mathcal{M}$:
\begin{equation}
\small
    (C_n,Q_n) = \mathop{{\rm argmin}}\limits_{(C',Q') \in \mathcal{M}}||h(C',Q'),\bm{k}_t(T_i)||.
\label{eq:neg}
\end{equation}

\begin{figure}[t]
\centering
\includegraphics[width = 0.8\linewidth]{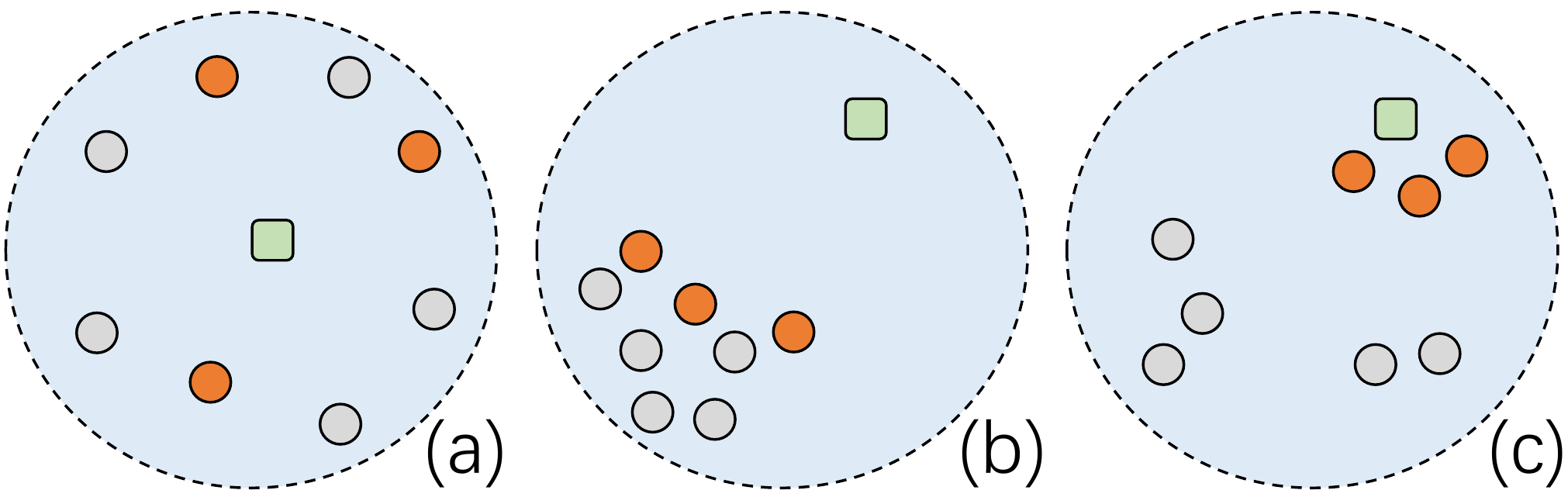}
\caption{Illustration of the diversity and locality property. (a) The diversity property distributes key vectors to the whole space. (b) The locality property cluster similar keys to facilitate knowledge sharing. (c) Diana aims to achieve a balance between diversity and locality}
\label{fig:balance}
\end{figure}

\textbf{Meta Prompt Keys}
help to combine these instance-level meta prompts $\{P_m^i\}_{i=1}^M$ to produce $P_m(C, Q)$. Specifically, for each input ($C, Q$), we select $M'$ meta prompt keys that are closest to its query vector $\bm{q}=h(C, Q)$. Then $P_m(C, Q)$ is obtained by concatenating these $M'$ meta prompts. Intuitively, the knowledge associated with ($C, Q, A$) is distributed in these $M'$ meta prompts.

When learning meta prompt keys, we expect the distribution of these keys to balance two properties: \emph{diversity} and \emph{locality} (Figure~\ref{fig:balance}). Specifically, the diversity property aims to distribute these keys to the whole vector space so that every meta prompt can be involved in the training process. The locality property aims to cluster similar meta prompts keys so that the knowledge of each sample can be better shared.
For each input $C$ and $Q$, we propose the following loss to enforce the above two properties:
\begin{equation}
\small
\begin{aligned}
    \mathcal{L}_{m} = \mathop{\sum}\limits_{i \in \mathcal{S}(C, Q)}{\rm  max}(0,||\bm{k}_m^{i}, h(C, Q)||-\eta)+\\\mathop{\sum}\limits_{i, j \in \mathcal{S}(C, Q)}{\rm max}(0,\gamma-||\bm{k}_m^{i},\bm{k}_m^{j}||)/{M'}^2,
\end{aligned}
\label{eq:meta_loss1}
\end{equation}
where $\mathcal{S}(C, Q)$ is the index set of these $M'$ meta prompt keys that are closest to $h(C, Q)$, $\eta$ and $\gamma$ are scalar hyper-parameters for the distance margin. Specifically, the first term in Eq. \ref{eq:meta_loss1} enforces the locality property by pulling these $M'$ meta prompt keys around the query vector. The second term enforces the diversity property by pushing these meta prompt keys away from each other to occupy the whole vector space.

Note that Eq.~\ref{eq:meta_loss1} only involves a single query $h(C, Q)$ from the current task. This may limit the learned meta prompt keys since samples from previously learned tasks are not considered. In this study, we extend Eq.~\ref{eq:meta_loss1} to better shape the distributions of meta prompt keys with the help of the memory buffer $\mathcal{M}$, in which samples from previously learned tasks are contained. Specifically, when learning the task $T_i$, we first calculate query vectors for samples in $\mathcal{M}$ and then group these query vectors into $B$ clusters (we set $B=5\times i$ in our experiments, where $i$ is the number of received tasks).
Centroids of these $B$ clusters are denoted as $\bm{c}_1, \cdots, \bm{c}_B$. 
For each sample ($C, Q$) from $\mathcal{M}$, the subsequent loss is optimized:
\begin{equation}
\small
\begin{aligned}
    \mathcal{L'}_m = \mathop{\sum}\limits_{i \in S(C, Q)}{\rm max}(0,||\bm{k}_m^{i},\bm{c}_k||-\eta),
\end{aligned}
\label{eq:meta_loss2}
\end{equation}
where $\bm{c}_k$ is the centroid to which ($C, Q$) belong. The above loss enforces the global diversity by scattering meta prompt keys to each centroid.

\subsection{Model Training} \label{sec:modeltraining}
\textbf{Scheduled Sampling of Task Prompts}
When training Diana, the task ID of each sample ($C, Q$) is given so that we can directly get the task prompt $P_t(T_i)$. However, naively using golden truth task IDs leads to an exposure bias issue, i.e., task IDs inferred in testing may not always be correct.

In this study, we introduce a scheduled sampling process to tackle the exposure bias issue. Specifically, for a given sample ($C, Q, A$) in the $k$-th training step, we toss a coin and use the golden truth task ID with probability $\epsilon_k$, or use the task ID inferred based on task prompt keys with probability $1-\epsilon_k$~\cite{bengio2015scheduled}. Note that when starting to learn each task, prompt keys are not well optimized, and thus the selected task ID is not accurate. Therefore, we set the value of $\epsilon_k$ to favor the golden truth task ID at the beginning (i.e., when $k$ is small) and gradually switch to the inferred task ID as the training proceeds (i.e., when $k$ is large), i.e., a linear decrement of $\epsilon_k$ is scheduled:
\begin{equation}
  \epsilon_k= {\rm max}(0,\alpha-k\beta),
  \label{eq:scheduled_sampling}
\end{equation}
in which $\alpha$ and $\beta$ are scalar hyper-parameters.

Note that LL models may encounter another source of exposure bias since we may receive inputs from unseen tasks in the testing phase. In this study, we use these $L$ extra prompts $\hat{P}_t(F_1), \cdots, \hat{P}_t(F_L)$ to explicitly model unseen tasks. Specifically, for each training sample ($C, Q, A$), we first determine its task format $F_j$ based on ($C, Q$), and allocate a small probability to use $\hat{P}_t(F_j)$ as its task prompt in $P(C, Q)$. In this way, we can capture general knowledge about all tasks for a given format in $\hat{P}_t(F_j)$ and expect the knowledge to facilitate handling unseen tasks.

\textbf{Train with LM Loss}
For each training sample ($C, Q, A$), we first construct the prompt $P(C, Q)$ using approaches introduced above, and then optimize $P(C, Q)$ together with the encoder-decoder model $g_\theta$ using the following LM loss:
\begin{equation}\label{eq:lm}
\mathcal{L}_{LM} = -log~g_\theta(A|[P(C, Q);C;Q]).
\end{equation}
The overall loss that we optimize for Diana is:
\begin{equation}
\mathcal{L} = \mathcal{L}_{m}+\mathcal{L'}_{m}+\mathcal{L}_{t}+\mathcal{L}_{LM}.
\end{equation}

After learning each task $T_i$, we select a small number of samples from $T_i$ based on the query vector of each sample to update the memory $\mathcal{M}$. This selection process aims to maintain diverse samples in $\mathcal{M}$. More details are in Appendix~\ref{app:mem}.

See summarized training process in Algorithm~\ref{alg:algorithm}.

\subsection{Model Inference}\label{sec:infer}
When testing, we determine the prompt $P(C,Q)$ for each input context $C$ and question $Q$, and use the learned model $g_\theta$ to predict the answer $A$.

\textbf{Adaptive Decision Boundaries (ADB)}
are used to select proper task prompts in the testing phase. Specifically, for each task $T_i$, a scalar boundary $\delta_i$ is constructed following the approach proposed by \citet{zhang2021deep}. An input ($C, Q$) is regarded as a sample from unseen tasks if its query vector $h(C, Q)$ falls outside the boundary of every task:
\begin{equation}
  || h(C, Q),\bm{k}_t(T_{i})||>\delta_i, \forall i \in [1, N].
\end{equation}
For samples from unseen tasks, we use the prompt $\hat{P}_t(F_j)$ as its task prompt in $P(C, Q)$, where $F_j$ is the format of ($C, Q$).

\textbf{Answer Prediction}
is performed with a greedy decoding process:
\begin{equation}
  A = \mathop{\rm arg max}_{A'} g_\theta(A' | [P(C, Q); C, Q]).
\end{equation}

\section{Experiments}
\subsection{Datasets}
We use two sets of tasks to evaluate Diana:

\textbf{1. decaNLP tasks:} We follow~\citet{sun2019lamol} to select 5 tasks from the decaNLP~\cite{decanlp} to train Diana.
These tasks cover 3 different formats: Span Extraction, Sequence Generation, and Text Classification. 
We also collect $N'=3$ additional tasks for each of these 3 format from decaNLP to serve as unseen tasks in the testing phase, i.e., our model is trained on $N=5$ seen tasks while tested on 8 tasks;

\textbf{2. QA tasks:} The second set focuses on question answering (QA) benchmarks.
Specifically, we use 8 QA datasets over 3 QA formats, i.e., Extractive QA, Abstractive QA and Multiple-Choice QA to train Diana.
We also collect $N'=3$ additional QA datasets for each of these three formats as unseen tasks, i.e., our model is trained on $N=8$ seen tasks while tested on 11 tasks.

Note that task IDs for all testing samples are not available in our experiments.
See Appendix~\ref{append:dataset},\ref{app:case} for more details of our dataset settings.

\subsection{Evaluation Metrics}
Individual tasks from above two task sets are evaluated following \citet{decanlp} and \citet{zhong2022proqa}, respectively (see Appendix \ref{append:dataset}).
To evaluate the LL performance of Diana, we build a performance matrix $R \in \mathbb{R}^{N \times (N+N')}$, where $R_{i,j}$ is the model performance on task $T_j$ after learning task $T_i$. The following LL metrics are computed:

\textbf{1. Average Performance} $A_N$ and $A_{N'}$ is defined as the average performance of the final model on $N$ seen tasks and $N'$ unseen tasks, respectively:
\begin{equation}
\small
    A_N = \frac{1}{N}\mathop{\sum}_{j=1}^N R_{N,j}, \quad A_{N'} = \frac{1}{N'}\mathop{\sum}_{j=N+1}^{N+N'} R_{N,j}.
\end{equation}

\textbf{2. Average Forget} $F_N$ is defined as the average performance decrease of each task after it is learned:
\begin{equation}
\small
    F_N = \frac{1}{N-1}\mathop{\sum}_{j=1}^{N-1} \mathop{{\rm max}}_{i \in \{1,\cdots,N-1\}} (R_{i,j} - R_{N,j}).
\end{equation}

In our experiments, we perform five runs with different random seeds and task orders. All reported metric scores are averages of these five runs. 
Ideally, we expect a strong LL model to yield high $A_N$ and $A_{N'}$ scores, and low $F_N$ scores.

\subsection{Implementation Details}
\label{sec:imp}
We use T5-base~\cite{raffel2020exploring} to initialize our encoder-decoder model, and set the lengths of soft prompts $P_g$, $P_f$, $P_t$, $P_m$ to 20, 40, 40, 20, respectively. We maintain totally $M=30$ meta prompts, and for each sample ($C, Q$) we choose $M'=5$ meta prompts to construct $P_m(C, Q)$. We use the AdamW~\cite{loshchilov2017decoupled} optimizer with a learning rate of 1e-4 and batch size of 64. Each task is trained for five epochs. We set $\eta = 0.15$ and $\gamma = 0.3$ in Eq.~\ref{eq:meta_loss1} and $\alpha=0.9$ and $\beta=3e-4$ in Eq.~\ref{eq:scheduled_sampling}. We maintain $50$ samples from each learned task in the memory $\mathcal{M}$. All experiments are performed on 4 V100 GPUs, and the computational cost of our model is analyzed in Appendix~\ref{app:com}. See more details in Appendix~\ref{app:de}.

\begin{table}[!t]
\centering
\small
\scalebox{0.9}{
    \setlength\tabcolsep{2.5pt} 
\begin{tabular}{c|l|c|cc|cc}
\toprule
    \multirow{2}{*}{\shortstack{\\Task ID\\in Test}} & \multirow{2}{*}{Methods} & \multirow{2}{*}{\shortstack{\\Buffer\\Size}} & \multicolumn{2}{c|}{QA Tasks} &\multicolumn{2}{c}{decaNLP Tasks} \\
    \cmidrule{4-7}
    & & & $A_N$ & $F_N$ & $A_N$ & $F_N$\\

\midrule
 \multirow{2}{*}{Yes} &ProQA  & 0  & 50.69 & 12.10 & 66.70 & 10.54 \\

& ProQA+ER & 50  &   54.00 & 7.27 & 71.26 & 5.33\\
\midrule
\multirow{12}{*}{No} & Finetune    & 0    & 46.81 &15.47  &57.92 &18.41 \\
&EWC       & 0   &47.81 &14.55  & 63.17  &13.58    \\
&FLCB      & 0   & 47.50 & 14.98 & 63.86 & 13.36 \\
&AdapterCL & 0   &48.08 &13.29 &64.25 &12.38\\
&L2P       & 0   & 48.15 & 13.89 & 63.76 & 13.47 \\
&DualPrompt& 0   & 48.54 & 13.66 & 64.47 & 12.49 \\
& ER       & 50  & 51.30 & 10.72 & 68.17 & 7.42\\
&DER++     & 50  & 52.01 & 10.05 & 69.10 & 6.86 \\
&AFPER     & 50  &52.69 & 9.28 &69.78 & 6.17 \\
\cmidrule{2-7}
&Diana w/o $\mathcal{M}$  & 0  & 50.30 & 12.68 & 66.14 & 10.61 \\
&Diana  & 50 &  \textbf{55.93} & \textbf{6.75} & \textbf{72.70} & \textbf{4.25} \\
\cmidrule{2-7}
& Multitask & - &59.23 & - &77.97 & - \\
\bottomrule
\end{tabular}}
\caption{Model performance on seen tasks. Best results (except the upper bound Multitask) are bolded. Our model Diana significantly outperforms other baselines on all metrics with $p$-value$<$0.05 ($t$-test).}
\label{tab:seen}
\end{table}

\subsection{Baselines}
We use the following competitive baselines covering all three types of LL models:

1. \emph{Regularization-based methods}: \textbf{EWC}~\cite{kirkpatrick2017overcoming} adopts the elastic weight consolidation approach to add regularization on parameter changes; \textbf{FLCB}~\cite{gao2022forget} uses knowledge learned from previous tasks to guide future task learning; 2. \emph{Rehearsal-based methods}: \textbf{ER}~\cite{chaudhry2019tiny} replays memory samples from previous tasks to consolidate learned knowledge; \textbf{DER++}~\cite{buzzega2020dark} augments ER with a $L_2$ loss on the soft labels; \textbf{AFPER}~\cite{mi-etal-2020-continual} combines ER with an adaptive elastic weight consolidation mechanism; 3. \emph{Architecture-based methods}: \textbf{AdapterCL}~\cite{madotto-etal-2021-continual} allocates separate adapters for different tasks; \textbf{L2P}~\cite{Wang_2022_CVPR} attaches a group of prompts on a pre-trained model to share fine-grained knowledge; \textbf{DualPrompt}~\cite{wang2022dualprompt} uses different prompts to encode task-invariant and task-specific knowledge; \textbf{ProQA}~\cite{zhong2022proqa} uses a unified structural prompt to implement LL models. Note that ProQA is designed for task incremental learning that requires accessing task IDs in the testing phase.

We combine ProQA and ER to implement a stronger baseline \textbf{ProQA+ER}, in which samples from previous tasks are replayed for the ProQA model, and we also implement a variant of Diana by removing the memory buffer \textbf{Diana w/o $\mathcal{M}$}. We further report the performance for sequentially fine-tuning the LL model on all tasks (\textbf{Finetune}) and multi-task learning (\textbf{Multitask}). Note that the performance of Multitask is generally regarded as the upper bound of LL models when only seen tasks are considered.\looseness=-1

All the above baselines are implemented following the same settings of our model, including using the same backbone PLM, prompt size, and memory size used for replay. Note that for the ProQA baseline, we follow its original setting to provide task IDs for testing samples when evaluating.\looseness=-1

\subsection{Experiment Results}
\paragraph{Results on Seen Tasks}
Table~\ref{tab:seen} shows the result on seen tasks from our two task sets. It can be seen that Diana outperforms all competitive baselines. Specifically, in the more general domain incremental learning scenario, i.e., when task IDs are unavailable in testing, Diana outperforms the best-performing baseline AFPER by a large margin. On QA tasks, Diana achieves 6.15\% relative improvement on the $A_N$ score and 27.26\% relative decrease on the $F_N$ score. Similar trend is also observed on decaNLP tasks. This means that Diana obtains higher performance with less forgetting in the LL process compared with other baselines.

We can also observe that: (1) Diana even outperforms the ProQA+ER baseline, which leaks task IDs in testing. This proves the superiority of our model design. (2) When task IDs are unavailable, Diana w/o $\mathcal{M}$ outperforms all baselines that do not use the memory buffer. This demonstrates that Diana's hierarchical prompts help to improve the LL performance even without the memory buffer.

\paragraph{Results on Unseen Tasks}
Table~\ref{tab:unseen} shows the result on unseen tasks from our two task sets. Note that we cannot compute the average forget score for unseen tasks since these tasks are never learned. Diana yields the best performances on all settings. It also achieves a relative improvement of 9.49\% and 11.04\% on the $A_{N'}$ score compared with the best baseline DER++ on these two task sets.

We can also observe that: (1) When $\mathcal{M}$ is unavailable, models that share knowledge through fine-grained components (i.e., Diana and L2P) generally obtain high performance, and our model that allocates extra prompts for unseen tasks achieves the best performance. This validates our approach of using hierarchical prompts to explicitly model unseen tasks. (2) It is interesting to see that Diana even outperforms Multitask, which is usually regarded as the upper bound of traditional LL models when only seen tasks are considered. This indicates that traditional LL models have limited generalization ability to unseen tasks and it also proves that our model is effective in modeling unseen tasks.

See Appendix~\ref{sec:experiments} for detailed experimental results of all tasks.

\begin{table}[!t]
\centering
\small
\scalebox{0.9}{
    \setlength\tabcolsep{3pt} 
\begin{tabular}{c|l|c|cc}
\toprule
 \multirow{2}{*}{\shortstack{\\Task ID\\in Test}}& \multirow{2}{*}{Methods}    & \multirow{2}{*}{\shortstack{\\Buffer\\Size}}  &  \multicolumn{2}{c}{$A_{N'}$} \\
\cmidrule{4-5}
& & & QA Tasks & decaNLP Tasks\\
\midrule
\multirow{2}{*}{Yes} & ProQA     & 0     & 35.85 & 30.08 \\
& ProQA+ER  & 50    & 38.00 & 30.92\\
\midrule
\multirow{12}{*}{No}& Finetune  & 0     & 35.51 & 28.08\\
& EWC       & 0     & 36.07  & 29.76  \\
& FLCB      & 0     & 36.68   & 31.17 \\
& AdapterCL & 0     & 36.84   & 30.32\\
& L2P       & 0     & 37.60  & 31.19 \\
& DualPrompt& 0     & 36.66 & 29.71  \\
& ER        & 50    & 37.80 & 30.05 \\
& DER++     & 50    & 38.47  & 31.24 \\
& AFPER     & 50    & 36.79  & 30.22\\
\cmidrule{2-5}
& Diana w/o $\mathcal{M}$ & 0  & 39.22 & 33.19 \\
& Diana  & 50 &  \textbf{42.12}  & \textbf{34.69}\\
\cmidrule{2-5}
& Multitask & -  & 40.62  & 32.72 \\
\bottomrule

\end{tabular}}
\caption{Model performance on unseen tasks. Best results are bolded. Diana significantly outperforms other baselines on all metrics with $p$-value$<$0.05 ($t$-test).}
\label{tab:unseen}
\end{table}

\subsection{Ablation Studies}
We conduct ablation studies on different components of Diana. Specifically, three types of variants are implemented:

1. Each of these four prompt types is ablated: \textbf{w/o general prompt}, \textbf{w/o format prompt}, \textbf{w/o task prompt}, \textbf{w/o meta prompt}.

2. Schemes to enhance task prompts are ablated: 
\textbf{w/o Sched. Sampling} removes the scheduled sampling scheme and only uses the ground truth task IDs in training;
\textbf{w/o G.T. Identity} is similar to the above variant. Instead, it only uses predicted task IDs in training;
\textbf{w/o Neg. Samples} only uses positive samples to train task prompt keys, i.e., the second term in Eq.~\ref{eq:metric_learning} is removed;
\textbf{w/o ADB} uses fixed decision boundaries instead of ADBs to detect unseen tasks.

3. Schemes to enhance meta prompts are ablated:
\textbf{w/o Sample Dive.} does not enforce the diversity property of the meta prompt keys, i.e., the second term in Eq.~\ref{eq:meta_loss1} is removed;
\textbf{w/o Memory Dive.} does not use samples from previous tasks to enhance the diversity property, i.e., the loss $\mathcal{L}'_m$ (Eq.~\ref{eq:meta_loss2}) is removed;
\textbf{w/o Loc.} does not enforce the locality property of the meta prompt keys, i.e., the first term in Eq.~\ref{eq:meta_loss1} is removed;
\textbf{w/o Cluster} does not cluster samples in $\mathcal{M}$, i.e., $\bm{c}_k$ in Eq.~\ref{eq:meta_loss2} is replaced with the query vector of each sample from $\mathcal{M}$.

\begin{table}[!t]
\centering
\small
\scalebox{0.9}{
\begin{tabular}{c|l|c|c|c}
\toprule
     Categories & Variants & $A_N$ & $F_N$ & $A_{N'}$\\
\midrule
\multirow{4}{*}{\shortstack{Prompt\\Types}} 
& w/o General Prompt & 55.47 & 6.93 & 40.74 \\
& w/o Format Prompt  & 55.11 & 7.03 & 40.59 \\
& w/o Task Prompt    & 53.87     & 8.50  & 39.66         \\
& w/o Meta Prompt    & 53.46      & 8.56  & 40.04       \\
\midrule
\multirow{4}{*}{\shortstack{Task\\prompt}} 
& w/o Sched. Sampling & 55.15 & 7.43 & 42.00 \\
& w/o G.T. Identity & 54.16 & 7.61 & 41.27 \\
& w/o Neg. Samples & 54.97 & 7.66 & 41.78 \\
& w/o ADB & 55.48 & 6.98 & 41.01 \\
\midrule
\multirow{3}{*}{\shortstack{Meta\\prompt}}
& w/o Sample Dive.   & 55.24      & 6.91  & 41.23        \\
& w/o Memory Dive.   & 55.02 & 7.41 & 41.48 \\
& w/o Loc.           & 54.70 & 7.54 & 41.16 \\
& w/o Cluster        & 55.46& 6.99 & 41.51\\
\midrule
\multicolumn{2}{c|}{Diana} & \textbf{55.93} & \textbf{6.75} & \textbf{42.12} \\
\bottomrule
\end{tabular}
}
\caption{Ablation studies of model components and training strategies on QA tasks. Each result is an average of 5 random runs.}
\label{tab:ablation}
\end{table}

Table \ref{tab:ablation} shows the performance of the above variants on QA tasks. It can be observed that Diana outperforms all the above variants. We can also see that: (1) ``w/o Meta Prompt'' lowers the LL performance by a large margin. This indicates that these fine-grained meta prompts are important in building lifelong models. (2) The scheduled sampling scheme helps to learn better task prompts and thus improves the LL performance. (3) ADB improves model performance on unseen tasks (i.e., $A_{N'}$) by a large margin. (4) Enforcing the diversity property of meta prompt keys is important to obtain good key representations and facilitates the learning of each task.

\subsection{More Analysis}
\subsubsection{Task ID Detection Performance}
Diana needs to detect task IDs of input samples when determining the task prompt to be used. To verify the performance of the task ID detector implemented in Diana (Section \ref{sec:keyspace} and \ref{sec:infer}), we compare the approach used in Diana with other task ID detectors: (1) Perplexity-based detector implemented in baseline ``AdapterCL'' determines the task IDs based on the perplexity of the PLM when different adapter modules are activated. (2) Distance-based detector implemented in our variant ``w/o Neg. Samples'' determines the task identity based on the distance between each key and query vectors. (3) Advanced distance-based detector implemented in our variant ``w/o ADB'' utilizes negative samples based on the above detector. Note that we do not apply ADB in the above two distance-based detectors. On our testing data, the above three approaches achieve a task ID detection accuracy of 59.84\%, 52.72\%, and 63.43\%, respectively, while Diana reaches a task ID detection accuracy of 66.97\%. This verifies the effectiveness of our approaches to optimize task prompt keys in detecting task IDs. More detailed comparisons of these task ID detectors can be found in Appendix~\ref{app:tid}.

\begin{table}[t]
\centering
\small
\scalebox{0.9}{
\begin{tabular}{c|l|cccc}
\toprule
    Criteria & Models  & $Z$=2 & $Z$=3 & $Z$=5 & $Z$=10 \\
\midrule
 \multirow{3}{*}{Locality} & w/o Sample Dive.  & 0.73 & 0.72  & \textbf{0.70}   & 0.48\\
& w/o Memory Dive. & 0.74  & 0.72 & 0.69 & 0.63 \\
& Diana  & \textbf{0.74}  & \textbf{0.73}  & \textbf{0.70}   & \textbf{0.66}   \\
\midrule
 \multirow{3}{*}{Diversity} & w/o Sample Dive.  & 0.63   & 0.61  & 0.59   & 0.40  \\
& w/o Memory Dive. & \textbf{1.00} & 0.89 &0.77 &0.53 \\
& Diana  & \textbf{1.00}  & \textbf{0.96}  & \textbf{0.89}   & \textbf{0.63}   \\
\bottomrule
\end{tabular}
}
\caption{Quantitative analysis of the locality and diversity for meta prompt keys on QA tasks.}
\label{tab:quantitative}
\end{table}

\subsubsection{Distribution of Meta Prompt Keys}
We also analyze the distribution of meta prompt keys $\mathcal{K} = \{\bm{k}_m^j\}_{j=1}^M$ constructed in Diana, which are expected to balance the locality and diversity property.
Specifically, we introduce two metrics to quantify these two properties.
For the diversity property, we follow \citet{coverage} to measure whether these meta prompt keys cover the whole vector space:
\begin{equation}
  \small
  Diversity = |\mathop{\cup}\limits_{j=1}^{M}\mathcal{N}_Z(\bm{k}_{m}^j, \mathcal{M})| / {(Z \cdot M)},
\end{equation}
where $\mathcal{N}_Z(\bm{k}_{m}^j, \mathcal{M})$ represents the set of top-$Z$ nearest samples in $\mathcal{M}$ around $\bm{k}_{m}^j$, and $|\cdot|$ returns the sample count of a set. High diversity scores are received if we can scatter meta prompt keys near every query vector from $\mathcal{M}$.
For the locality property, we follow \citet{scellato2010distance} to measure whether there are keys clustered around each query vector $\bm{q}$ in $\mathcal{M}$:
\begin{equation}
  \small
  Locality = \mathop{\sum}\limits_{\bm{q} \in \mathcal{M}}\mathop{\sum}\limits_{\bm{k} \in \mathcal{N}_Z(\bm{q}, \mathcal{K})}{(1-||\bm{q}, \bm{k}}||) / (Z \cdot |\mathcal{M}|).
\end{equation}
High locality scores are received if meta prompt keys in $\mathcal{K}$ are tightly clustered.

On the QA tasks, we compare the above two metrics between Diana and our ablation variants for meta prompts under different values of $Z$.
As can be seen from Table~\ref{tab:quantitative}, the strategies we introduced in Diana (Section~\ref{sec:keyspace}) help to enforce the locality and diversity properties of meta prompt keys.\looseness=-1

\section{Conclusion}
We propose Diana, a novel LL model for the domain incremental learning scenario. Diana converts different NLP tasks into a unified sequence generation format and uses a prompt-enhanced PLM to learn these tasks. We introduce four types of hierarchically organized prompts in Diana to capture knowledge in different granularities. 
These prompts are shared between different scopes of samples and are dynamically combined based on a set of key vectors. The space of key vectors is learned with several distance-based regularization terms. Dedicated components are also allocated in Diana to model samples from unseen tasks. Experiments and empirical analysis on two sets of tasks show that Diana outperforms SOTA LL models, especially in handling samples from unseen tasks.

\section*{Limitations}
One major limitation of this study is its input modality. Specifically, our model is limited to textual inputs and ignores other modalities (e.g., vision and audio). Open and domain incremental lifelong learning across modalities is more realistic and challenging. Fortunately, we can obtain robust features of different modalities via multi-modal pre-training models~\cite{xu-etal-2021-layoutlmv2,DBLP:journals/corr/abs-2103-06561}. For future work, we will try to tackle multi-modal tasks in an open (including out of distribution data~\citep{lang-etal-2022-estimating,lang2023out,lang2023survey}) and domain incremental lifelong learning scenario with better approaches.
\section*{Ethics Statement}
This work does not raise any direct ethical issues.
In the proposed work, we seek to develop a model for domain incremental lifelong learning in an open world, and we believe this work leads to intellectual merits that benefit from a realistic and efficient lifelong learning model. All experiments are conducted on open datasets.

\bibliography{custom,anthology}
\bibliographystyle{acl_natbib}

\appendix

\section{More Implementation Details}\label{app:de}
We use T5-base~\cite{raffel2020exploring} to initialize our encoder-decoder model (12 layers, 768 dimensional hidden size, and 12 attention heads), and set the lengths of soft prompts $P_g$,$P_f$,$P_t$,$P_m$ to 20, 40, 40, 20, respectively. We use a fixed T5-base encoder with an average pooling layer to obtain the query vector. We maintain a pool of $M=30$ meta prompts, and for each sample ($C, Q$) we choose $M'=5$ meta prompts to construct $P_m(C, Q)$. We use the AdamW~\cite{loshchilov2017decoupled} optimizer for training. All hyper-parameters are tuned according to the average score on validation datasets of NarQA, RACE, OBQA, SIQA and Dream. We tried epoch number of $\{2,3,4,5,6,7,8\}$ and learning rate of $\{1e-5,5e-5,1e-4,5e-4,1e-3\}$. We finally set the learning rate to 1e-4 and the number of training epochs to 5. We set $\eta = 0.15$ and $\gamma = 0.3$ in Eq.~\ref{eq:meta_loss1} and $\alpha=0.9$ and $\beta=3e-4$ in Eq.~\ref{eq:scheduled_sampling}. For $\eta$ and $\gamma$, we have a grid search between 0 and 0.5 with an interval of 0.05. For $\alpha$ and $\beta$, $\alpha$ is searched among $\{0.9,0.7,0.5\}$, while $\beta$ is searched among $\{1e-5,3e-5,1e-4,3e-4,1e-3\}$. All experiments are performed on 4 V100 GPUs (32GB). The batch size is set to 64. In each set of tasks, We perform 5 runs with different task orders by setting the random seed to $\{42,43,44,45,46\}$ respectively. In this way, we report the average score of each method. Note that we only use the random seed $42$ for tuning hyper-parameters.

In order to train extra task prompts $\{\hat{P}_t(F_1),\cdots,\hat{P}_t(F_L)\}$ for unseen tasks, we allocate a small probability $\omega = 5\%$ for each training sample ($C, Q, A$) to use $\hat{P}_t(F_j)$ as its task prompt in $P(C, Q)$, where $F_j$ is the task format of ($C, Q, A$). To implement variant ``w/o ADB'' for ablation study, we use a fixed decision boundary instead of ADB. If for any task $T_i$, the distance $||h(C,Q),\bm{k}_t(T_i)||>0.35$, we regard the sample is from unseen tasks.

The adaptive decision boundary for each task is determined following the approach proposed by~\citet{zhang2021deep}. We use AdamW optimizer with a learning rate of 0.02 to learn each decision boundary.
To obtain the ROUGE-L score, we use the NLTK package for sentence tokenization, and python rouge-score package for evaluation.

\section{Memory Update}\label{app:mem}
After learning task $T_i$, we select $E$ diverse samples (we set $E=50$ in our experiments) from $T_i$ to update the memory $\mathcal{M}$ based on the query vector of each sample. Specifically, our selection criteria are built based on the distance of these prompt keys and query vectors. For each meta prompt key $\bm{k}_m^j$ ($j=1,\cdots,M$), we select top-$\lceil \frac{E}{M} \rceil$ samples ($\lceil\cdot\rceil$ is the ceiling function), whose query vectors are closest to $\bm{k}_m^j$. After accumulating $M\lceil \frac{E}{M}\rceil$ memory samples selected by $M$ meta prompt keys, we rank these samples based on their distance to the corresponding meta prompt keys, and choose top-$E$ samples with the smallest distance to be fed into $\mathcal{M}$. In this way, the memory $\mathcal{M}$ we constructed can expand to the whole space of prompt keys.

Note that, the memory buffer $\mathcal{M}$ is optional in Diana. Without $\mathcal{M}$, the loss in Eq.~\ref{eq:meta_loss2} is not optimized, and the second term in Eq.~\ref{eq:metric_learning} is removed.

\section{Detailed Dataset Setting and Evaluation Metrics}\label{append:dataset}
For the decaNLP task set, 8 benchmarks over 3 formats are covered, i.e., (1) \emph{Span Extraction}, including \textbf{SQuAD}~\cite{rajpurkar-etal-2016-squad}, \textbf{QA-ZRE}~\cite{levy-etal-2017-zero}, \textbf{QA-SRL}~\cite{he-etal-2015-question}; (2) \emph{Sequence Generation}, including \textbf{WOZ}~\cite{wen-etal-2017-network}, \textbf{WikiSQL}~\cite{zhong2017seq2sql}, \textbf{CNN/DM}~\cite{NIPS2015_afdec700}; (3) \emph{Text Classification}, including \textbf{SST}~\cite{socher-etal-2013-recursive} and M\textbf{NLI}~\cite{williams-etal-2018-broad}. For the QA task set, 11 QA benchmarks over 3 QA formats are covered, i.e., : (1) \emph{Extractive} QA, including \textbf{SQuAD}~\cite{rajpurkar-etal-2016-squad}, \textbf{NewsQA}~\cite{trischler-etal-2017-newsqa}, and \textbf{Quoref}~\cite{dasigi-etal-2019-quoref}; (2) \emph{Abstractive} QA, including \textbf{NarQA}~\cite{TACL1197}, \textbf{NQOpen}~\cite{TACL1455}, and \textbf{Drop}~\cite{dua-etal-2019-drop}; (3) \emph{Multiple-Choice} QA, including \textbf{RACE}~\cite{lai-etal-2017-race}, \textbf{OBQA}~\cite{mihaylov-etal-2018-suit}, \textbf{MCTest}~\cite{richardson-etal-2013-mctest}, \textbf{SIQA}~\cite{sap-etal-2019-social}, and \textbf{Dream}~\cite{sun2019dream}. The statistics of the above 
datasets are summarized in Table~\ref{tab:dataset}. We follow the pre-process scheme released by \citet{khashabi-etal-2020-unifiedqa} to tackle these datasets. Some of these datasets do not contain a validation set, thus we only use the validation sets of NarQA, RACE, OBQA, SIQA and Dream in the QA task set to search hyper-parameters.

The evaluation for each single task follows~\citet{decanlp,zhong-etal-2022-proqa}. Among the decaNLP tasks, we compute F1 score for QA-SRL and QA-ZRE, Exact Match (EM) score for SQuAD, MNLI and SST, ROUGE-L for CNN/DM. For WOZ, we adopt turn-based dialogue state
exact match (dsEM). For WikiSQL, we use exact match of logical forms (lfEM). For the QA task set, we compute the accuracy of option selection for all Multi-Choice QA tasks and use EM score for all Extractive QA tasks. Among Abstractive QA tasks, we use F1 score for Drop and NQOpen, and ROUGE-L~\cite{lin-2004-rouge} for NarQA. 

\begin{table}[!t]
\centering
\small
\scalebox{0.8}{
\begin{tabular}{c|c|c|c|c}
\toprule
Task set   & Dataset  & Train set size & Val set size & Test set size \\
\midrule
\multirow{8}{*}{decaNLP} & SQuAD & 87k  & - & 10k \\
 & QA-ZRE & - & - & 12k \\
 & QA-SRL & 6.4k & - & 2.2k \\
 & WikiSQL & 56k & - & 15k \\
 & WOZ & 2.5k & - & 1.6k \\
 & CNN/DM & -  & - & 11k \\
 & SST & 6.9k & - & 1.8k \\
 & MNLI & - & - & 20k \\
\midrule
 \multirow{11}{*}{QA} & SQuAD & 87k  & - & 10k \\
       & NewsQA    & 76k  & - & 4.3k   \\
      & Quoref     & - & -  & 2.7k  \\
   & NarQA   & 65k & 6.9k & 21k  \\
  & NQOpen & 9.6k & - & 10k\\
  & Drop & - & - & 9.5k\\
  & RACE      & 87k     & 4.8k & 4.9k \\
 & OBQA     & 4.9k  & 500 & 500 \\
   & MCTest     & 1.4k   & - & 320 \\
  & SIQA     & 33k & 1.9k & 2.2k\\
    & Dream    & -   & 2.0k & 2.0k \\
\bottomrule
\end{tabular}}
\caption{Dataset Statistics of the decaNLP task set and the QA task set.}
\label{tab:dataset}
\end{table}

\section{Detailed Experimental Results}\label{sec:experiments}
We provide the detailed performance of Diana under each single task compared with competitive baselines. The results under five seen tasks of the decaNLP task set, and eight seen tasks of the QA task set are shown in Table~\ref{tab:deca-seen-detail} and Table~\ref{tab:qa-seen-detail}. The results of unseen tasks for the decaNLP task set and the QA task set are shown in Table~\ref{tab:deca-unseen-detail} and Table~\ref{tab:qa-unseen-detail}.

\begin{table*}[!t]
\centering
\small
\scalebox{0.9}{
\begin{tabular}{c|l|c|ccccc|c|c}
\toprule
    Task-ID  & \multirow{2}{*}{Methods}       & Buffer &
    \multicolumn{5}{c|}{$R_{N,j}$} &
    \multirow{2}{*}{$A_N$} &
    \multirow{2}{*}{$F_N$} \\
    \cmidrule{4-8}
 in Test  & & Size & SQuAD &WikiSQL &SST &QA-SRL &WOZ & \\

\midrule
 \multirow{2}{*}{Available} &ProQA  & 0  & 71.09 & 37.39 & 92.16 & 75.68 & 57.17 & 66.70 & 10.54 \\

& ProQA+ER & 50 & 75.57 & 50.98 & 91.67 & 76.74 & 61.33 & 71.26 & 5.33 \\
\midrule
\multirow{12}{*}{\shortstack{Unavailable}} & Finetune    & 0    & 68.09 & 19.70 & 90.45 & 69.43 & 41.91 &57.92 &18.41            \\
&EWC  & 0   & 70.57 & 35.97 & 89.79 & 71.19 & 48.34 & 63.17  &13.58    \\
&FLCB  & 0 & 70.96 & 33.35 & 90.03 & 74.71 & 50.23  & 63.86 & 13.36 \\
&AdapterCL & 0 & 71.82 & 35.14 & 90.95 & 72.83 & 50.53 &64.25 &12.38\\
&L2P & 0 & 70.18 & 34.62 & 90.39 & 72.57 &  51.02  & 63.76 & 13.47 \\
&DualPrompt & 0 & 70.99 & 35.33 & 90.91 & 73.92 & 51.18 & 64.47 & 12.49 \\
& ER & 50 & 73.65 & 47.96 & 92.20 & 74.17 & 52.88  & 68.17 & 7.42\\
&DER++    & 50  & 74.18 & 49.27 & 92.34 & 75.11 & 54.61 & 69.10 & 6.86 \\
&AFPER & 50 & 75.27 & 48.90 & 91.56 & 76.34 & 56.82 &69.78 & 6.17 \\
\cmidrule(r){2-10}
&Diana w/o $\mathcal{M}$  & 0 & 71.94 & 36.25 & 91.03 & 74.59 & 56.90 & 66.14 & 10.61 \\
&Diana  & 50 & \textbf{76.93} & \textbf{51.09} & \textbf{92.74} & \textbf{77.69} & \textbf{65.06} & \textbf{72.70} & \textbf{4.25} \\
\cmidrule(r){2-10}
& Multitask & - & 79.68 & 53.65 & 93.59 & 80.38 & 82.57 &77.97 & - \\
\bottomrule
\end{tabular}}
\caption{Model performance on seen tasks in decaNLP. Best results (except the upper bound Multitask) are bold. Our model Diana significantly outperforms other baselines on all metrics with $p$-value$<$0.05 ($t$-test).}
\label{tab:deca-seen-detail}
\end{table*}

\begin{table*}[!t]
\centering
\small
\scalebox{0.8}{
\begin{tabular}{c|l|c|cccccccc|c|c}
\toprule
    Task-ID  & \multirow{2}{*}{Methods}       & Buffer &
    \multicolumn{8}{c|}{$R_{N,j}$} &
    \multirow{2}{*}{$A_N$} &
    \multirow{2}{*}{$F_N$} \\
    \cmidrule{4-11}
 in Test  & & Size & SQuAD & NewsQA & NarQA & NQOpen & RACE & OBQA & MCTest & SIQA  &\\

\midrule
 \multirow{2}{*}{Available} &ProQA  & 0  & 67.66    & 38.73     & 37.96     & 37.72     & 53.75     & 43.73     & 68.27     & 57.73 & 50.69 & 12.10 \\

& ProQA+ER & 50  &  71.20    & 40.17     & 41.94     & 39.00     & 57.09     & 47.00     & 77.94     & 57.67 & 54.00 & 7.27 \\
\midrule
\multirow{12}{*}{\shortstack{Unavailable}} & Finetune    & 0    & 57.58      & 35.84  & 33.74 & 34.49 & 50.28 & 42.20 & 65.67 & 54.72 &46.81 &15.47            \\
&EWC     & 0    & 59.84       & 36.44     & 34.88 & 35.14      & 50.54     & 43.43  & 66.52 & 55.68  &47.81 &14.55    \\
&FLCB  & 0 & 58.73     & 36.97     & 34.27     & 34.90     & 51.63     & 41.53     & 66.60     & 55.39 & 47.50 & 14.98 \\
&AdapterCL & 0  & 59.64     & 37.31    & 37.42     & 36.70    & 49.57     & 41.80     & 66.67    & 55.54  &48.08 &13.29\\
&L2P & 0 & 62.98 & 36.23 & 35.79 & 36.49 & 49.00 & 41.93 & 66.98 & 55.77 & 48.15 & 13.89 \\
&DualPrompt & 0  & 62.60 & 36.36 & 34.35 & 36.53 & 52.10 & 42.67 & 67.57 & 56.26 & 48.54 & 13.66 \\
& ER & 50 & 65.08 & 38.72 & 39.07 & 36.48 & 55.90 & 43.53 & 74.31 & 57.29 & 51.30 & 10.72\\
&DER++    & 50   & 67.08     & 39.03    & 39.91    & 36.93     & 56.42     & 44.13     & 74.77     & 57.77 & 52.01 & 10.05 \\
&AFPER & 50 & 68.14    & 40.79     & 40.16    & 38.89    & 55.08     & 46.60     & 75.33     & 56.52 &52.69 & 9.28 \\
\cmidrule(r){2-13}
&Diana w/o $\mathcal{M}$  & 0 & 65.51     & 37.78     & 37.35     & 37.41     & 54.14     & 46.27     & 68.50     & 57.41 & 50.30 & 12.68 \\
&Diana  & 50 & \textbf{74.44}    & \textbf{42.91}     & \textbf{43.16}     & \textbf{40.05}     & \textbf{59.08}     & \textbf{48.47}     & \textbf{78.44}     & \textbf{60.92} & \textbf{55.93} & \textbf{6.75} \\
\cmidrule(r){2-13}
& Multitask & - & 80.22 & 44.74 & 47.30 & 41.72 & 64.05  & 51.00 & 83.44 &61.41 &59.23 & - \\
\bottomrule
\end{tabular}}
\caption{Model performance on seen QA tasks. Best results (except the upper bound Multitask) are bold. Our model Diana significantly outperforms other baselines on all metrics with $p$-value$<$0.05 ($t$-test).}
\label{tab:qa-seen-detail}
\end{table*}

\begin{table}[!t]
\centering
\small
\scalebox{0.9}{
\begin{tabular}{l|c|ccc|c}
\toprule
\multirow{2}{*}{Methods}         & Buffer & \multicolumn{3}{c|}{$R_{N,j}$}  & \multirow{2}{*}{$A_{N'}$} \\
\cmidrule{3-5}
 & Size & CNN/DM & QA-ZRE & MNLI & \\
\midrule
ProQA  & 0  & 13.25     & 37.58     & 39.42  & 30.08  \\
ProQA+ER & 50  & 14.18     & 38.42     & 40.17 & 30.92\\
Finetune & 0 & 10.61    & 36.50    & 37.12 & 28.08\\
EWC     & 0    & 11.78      & 37.62     & 39.88  & 29.76   \\
FLCB  & 0 &  12.98    & 40.02    & 40.52 & 31.17  \\
AdapterCL & 0  & 13.23     & 37.88     & 39.84  & 30.32 \\
L2P & 0 & 13.09 & 40.16  & 40.31 & 31.19 \\
DualPrompt & 0 & 12.92 & 37.04 & 39.18 & 29.71 \\
ER & 50 & 13.04 & 38.06 & 39.04 & 30.05 \\
DER++    & 50   & 14.67     & 39.74     & 39.32  & 31.24  \\
AFPER & 50 & 12.14     &  38.66   & 39.85   & 30.22  \\
\midrule
Diana w/o $\mathcal{M}$ & 0 & 14.94 & 43.95 & 40.69 & 33.19 \\
Diana  & 50 & \textbf{15.80}     & \textbf{44.74}     & \textbf{43.53}  & \textbf{34.69}  \\
\midrule
Multitask & - & 15.98 & 42.12 & 40.07  & 32.72 \\
\bottomrule

\end{tabular}}
\caption{Model performance on unseen tasks in decaNLP. Best results (except Multitask) are bold. Diana significantly outperforms other baselines on all metrics with $p$-value$<$0.05 ($t$-test).}
\label{tab:deca-unseen-detail}
\end{table}

\begin{table}[!t]
\centering
\small
\scalebox{0.9}{
\begin{tabular}{l|c|ccc|c}
\toprule
\multirow{2}{*}{Methods}         & Buffer & \multicolumn{3}{c|}{$R_{N,j}$}  & \multirow{2}{*}{$A_{N'}$} \\
\cmidrule{3-5}
 & Size & Quoref & Drop & Dream & \\
\midrule
ProQA  & 0  & 33.40     & 18.29     & 55.85  & 35.85  \\
ProQA+ER & 50  & 35.87     & 19.78     & 58.35 & 38.00\\
Finetune & 0 & 33.08.    & 18.10    & 55.36 & 35.51\\
EWC     & 0    & 33.43       & 18.14     & 56.65  & 36.07   \\
FLCB  & 0 & 34.85     & 18.31    & 56.88 & 36.68   \\
AdapterCL & 0  & 35.47     & 17.83     & 57.21  & 36.84  \\
L2P & 0 & 36.22 & 19.18  & 57.40 & 37.60 \\
DualPrompt & 0 & 35.22 & 18.52 & 56.25 & 36.66 \\
ER & 50 & 35.14 & 18.56 & 59.71 & 37.80 \\
DER++    & 50   & 36.15     & 19.08     & 60.17  & 38.47  \\
AFPER & 50 & 35.26     & 18.83    & 56.29   & 36.79  \\
\midrule
Diana w/o $\mathcal{M}$  & 0 & 37.95     & 20.32    & 59.39   & 39.22 \\
Diana  & 50 & \textbf{40.42}     & \textbf{22.91}     & \textbf{63.03}  & \textbf{42.12}  \\
\midrule
Multitask & - & 36.27 & 22.99 & 62.60  & 40.62 \\
\bottomrule

\end{tabular}}
\caption{Model performance on unseen QA tasks. Best results (except Multitask) are bold. Diana significantly outperforms other baselines on all metrics with $p$-value$<$0.05 ($t$-test).}
\label{tab:qa-unseen-detail}
\end{table}

\section{More Analysis of Task Identity Detection Performance}\label{app:tid}
Architecture-based LL models need to detect task identities of input samples when these identities are unavailable in the testing phase. To verify the performance of the task identity detector implemented in Diana, we compare our approach with other task identity detectors: (1) Perplexity-based detector implemented in baseline ``AdapterCL'' determines the task identities based on the perplexity of the PLM when different adapter modules are activated. (2) Distance-based detector implemented in our variant ``w/o Neg. Samples'' determines the task identity based on the distance between each key and query vectors. (3) Advanced distance-based detector implemented in our variant ``w/o ADB'' utilizes negative samples based on the above detector. Note that we do not apply ADB in the above two distance-based detectors. 

The above approaches are trained and evaluated with the QA tasks under two scenarios: (1) In \textbf{Closed-world}: detectors are only required to detect samples from seen tasks. Note that in this setting, the Advanced distance-based detector used in ``w/o ADB'' is the same as the task identity detector implemented in Diana. (2) In \textbf{Open-world}: detectors are required to handle unseen task samples as well. When tested in the open-world scenario, these two distance-based detectors adopt a fixed decision boundary of 0.35 (see Appendix~\ref{app:de}). The perplexity-based detector adopts a perplexity threshold of 4, i.e., samples with a perplexity score above 4 are regarded as unseen task samples. This perplexity threshold is selected based on the model performance on the validation set. 

We report the task identity detection accuracy and Marco F1 scores for seen samples and unseen samples separately in Table~\ref{tab:match}. we can observe that: (1) The task identity detector used in Diana achieves the best performance in both scenarios. This proves the effectiveness of our task prompt keys in detecting task identities. (2) Negative samples used in Advanced distance-based detector significantly improve the task identity detection performance on seen tasks. (3) ADB is effective in improving the task identity detection performance on unseen tasks.

\begin{table*}[t]
\small
\centering
\scalebox{1.}{
\begin{tabular}{c|l|cccccc}
\toprule
    \multirow{2}{*}{Scenario} & \multirow{2}{*}{Methods}  & \multicolumn{2}{c}{Scores on Seen Tasks} & \multicolumn{2}{c}{Scores on Unseen Tasks} & \multicolumn{2}{c}{Overall Scores} \\
    \cmidrule{3-8}
     & & F1 & Accuracy & F1 & Accuracy & F1 & Accuracy \\
\midrule
 \multirow{3}{*}{Closed-world} &Perplexity-based  & 44.92 & 52.20  & - & - & 44.92 & 52.20\\

&Distance-based  & 43.18  & 63.34  & - & - & 43.18 & 63.34   \\
&Advanced distance-based  & \textbf{54.37}  & \textbf{75.35} & - & - & \textbf{54.37} & \textbf{75.15}   \\
\midrule
 \multirow{4}{*}{Open-world} &Perplexity-based  & 33.15  & 58.64 & 26.14   & \textbf{62.98} & 32.37   & 59.84\\

&Distance-based  &  38.51 & 50.53 & 21.98 & 58.48 & 36.67 & 52.72 \\
&Advanced distance-based  & 44.12   & 64.86 & 24.17   & 59.67 & 41.90   & 63.43   \\
&Diana & \textbf{47.06}   & \textbf{68.81} & \textbf{35.70}   & 62.16 & \textbf{45.80}   & \textbf{66.97}  \\
\bottomrule
\end{tabular}
}
\caption{Task identity detection performance of different models under the QA tasks.}
\label{tab:match}
\end{table*}

\section{More Analysis of Scheduled Sampling}
We perform a more detailed analysis of the scheduled sampling scheme introduced in Diana.
Specifically, in the ablation variant ``w/o G.T. Identity'', the model only uses predicted task identities in training.
This scheme helps to alleviate the discrepancy between training and testing with the cost of the model coverage speed. 
In the ablation variant ``w/o Sched. Sampling'', the model only uses golden truth task identities in the training process. This scheme leads to the discrepancy between training and testing. The above two schemes under-perform our model Diana. 

\begin{figure}[t]
\scalebox{0.85}{
\centering
\includegraphics[width=1.0\linewidth]{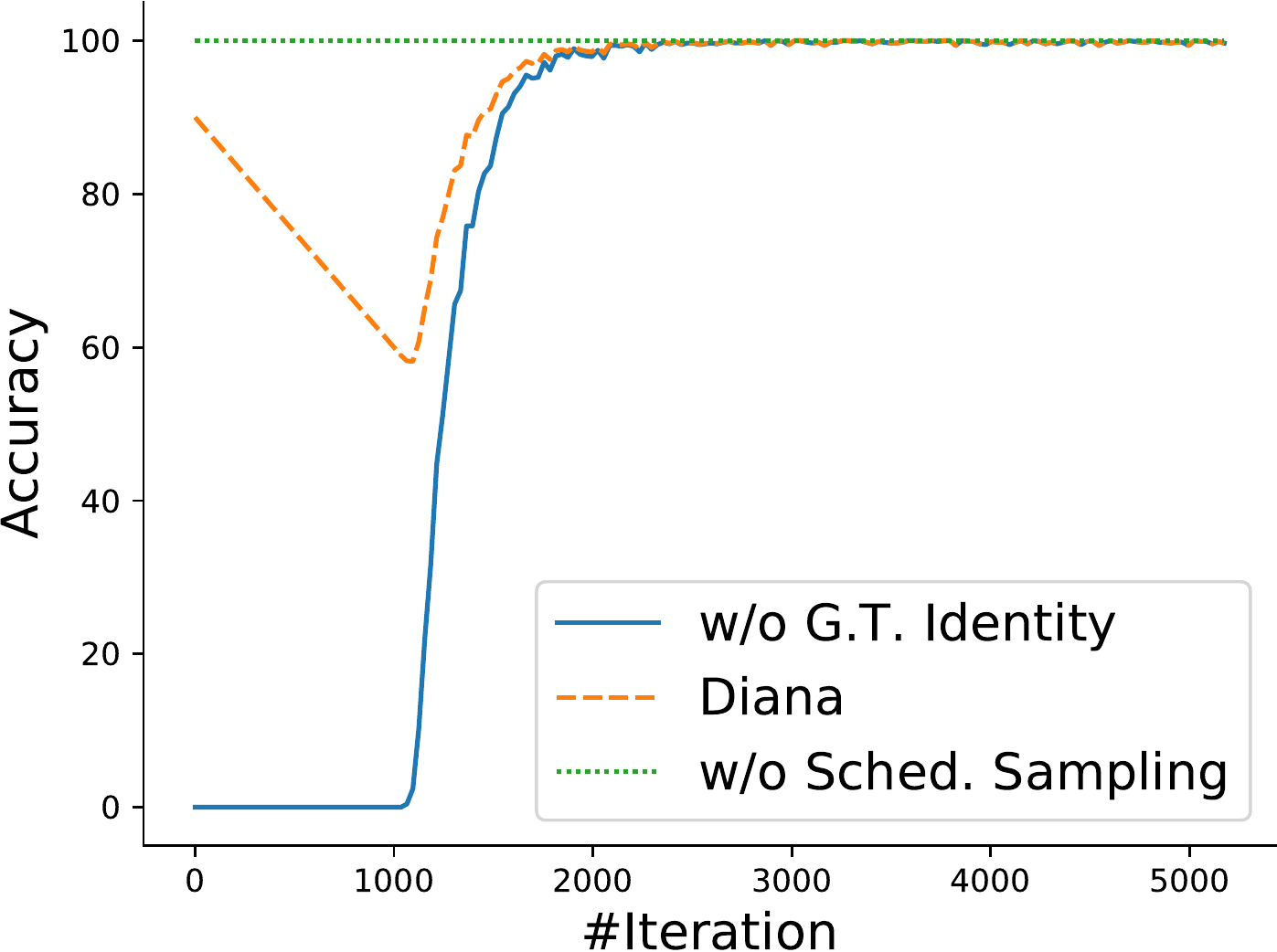}
}
\centering
\caption{The task identity detection accuracy for samples from the last task $T_N$ when learning $T_N$ of the QA task set.}
\label{fig:change}
\end{figure}

In this section, we analyze the task identity detection accuracy yield by the above schemes in Figure~\ref{fig:change} when learning the last task $T_N$ in the input task sequence of QA task set.
We can observe that the task identity detection accuracy achieved by ``w/o G.T. Identity'' is extremely low in earlier iterations,
which hinders task prompts from sharing task-specific knowledge in the early training stage. 
The scheduled sampling process introduced in Diana effectively compromises between detecting correct task identities and alleviating the train-test discrepancy,
and thus it results in the best LL performance among these variants.
Note that the task identity detection accuracy in ``w/o Sched. Sampling'' is almost zero in the first 1,000 iterations when learning task $T_N$.
This is because the task prompt keys for previous $N-1$ tasks are already well learned. 
The randomly initialized prompt key for task $T_N$ needs to be pulled to the query vector space before starting to be functional.


\section{More Analysis of Computational Cost}\label{app:com}
We analyze the computational cost of Diana when learning the QA tasks,
including the number of tunable parameters, time used for training and testing, and size of required memories retained from previous tasks. As indicated in Table~\ref{tab:cost}, Diana does not introduce too much computation overhead.
 
\begin{table}[!t]
\centering
\small
\scalebox{0.9}{
\begin{tabular}{l|c|c|c|c}
\toprule
 Methods & \shortstack{Tunable\\Parameters} & \shortstack{Memory\\Size} & \shortstack{Train Time\\Per Batch} & \shortstack{Test Time\\All Tasks} \\
\midrule
Lower Bound & 222.90M &  0 & 0.55 & 523 \\
EWC       & 222.90M    & 0  & 0.93 & 596   \\
FLCB     & 222.90M     & 0 & 0.59  & 591  \\
AdapterCL     & 262.25M & 0  & 0.73 & 5852  \\
L2P & 223.39M & 0 & 1.01 & 1013\\
DualPrompt & 223.17M & 0 & 0.93 & 1147\\
ER  & 222.90M      & 50     & 0.58 & 541 \\
DER++  & 222.90M     & 50  & 0.68 & 604 \\
AFPER   & 222.90M     & 50   & 0.95 & 630 \\
ProQA   & 223.43M     & 0 & 0.86 & 863\\
\midrule
Diana    & 223.84M     & 50   & 1.05 & 1108 \\
Diana w/o $\mathcal{M}$    & 223.84M     & 0   & 0.97 & 1123 \\
\bottomrule
\end{tabular}}
\caption{Computational cost of Diana and baselines for the QA task set. ``Train Time'' is the average time cost for each batch. ``Test Time'' is the total time cost to evaluate all 11 tasks. Both train and test times are in seconds.}
\label{tab:cost}
\end{table}

\section{Effect of PLM Size}
We evaluate Diana and the best-performing baseline DER++ on different sized PLM using QA datasets.
\begin{table}[!t]
\centering
\small
\scalebox{0.9}{
\begin{tabular}{c|l|c|c|c}
\toprule
     PLM Size & Method & $A_N$ & $F_N$ & $A_{N'}$\\
\midrule
\multirow{2}{*}{T5-small} 
& DER++ & 41.78 & 15.69 & 26.62 \\
& Diana  & \textbf{46.50} & \textbf{10.42} & \textbf{31.95} \\
\midrule
\multirow{2}{*}{T5-base} 
& DER++ & 52.01 & 10.05 & 38.47 \\
& Diana  & \textbf{55.93} & \textbf{6.75} & \textbf{42.12} \\
\midrule
\multirow{2}{*}{T5-large} 
& DER++ & 59.97 & 9.50 & 46.71 \\
& Diana  & \textbf{64.19} & \textbf{6.85} & \textbf{51.28} \\
\bottomrule
\end{tabular}
}
\caption{Performance with different sized PLMs on QA tasks.}
\label{tab:plmsize}
\end{table}
\begin{table}[!t]
\centering
\small
\scalebox{0.9}{
\begin{tabular}{c|c|c|c}
\toprule
Method & $A_N$ & $F_N$ & $A_{N'}$\\
\midrule
 Prompt tuning & 46.76 & \textbf{4.71} & 32.87 \\
 Full tuning  & \textbf{55.93} & 6.75 & \textbf{42.12} \\
\bottomrule
\end{tabular}
}
\caption{Performance with different training methods on QA tasks.}
\label{tab:training}
\end{table}
As shown in Table~\ref{tab:plmsize}, Diana obtains better performance with larger PLM size, and consistently outperforms the baseline.

\section{Analysis of Training Method}
During training, we follow a full tuning scheme that updates parameters of the backbone language models (T5) along with prompts. We also investigate the performance of prompt tuning, which fixes the backbone language model and only updates the prompts. As indicated in Table~\ref{tab:training}, prompt tuning dramatically degenerates the performance of Diana.

\section{Cases}\label{app:case}
We list some samples for tasks we modeled from the decaNLP task set and the QA task set respectively, shown in Table~\ref{tab:cases-deca} and Table~\ref{tab:cases}.

\section{Training Process}
Details about the training process of Diana are shown in Algorithm~\ref{alg:algorithm}.

\begin{table*}[!t]
\centering
\small
\scalebox{0.9}{
\begin{tabular}{c|c|l}
\toprule
Format    & Dataset  & Case \\
\midrule
\multirow{9}{*}{Span Extraction} & \multirow{3}{*}{SQuAD} & \textbf{Context}: (Private\_school) Private schooling in the United States has been... \\
& &\textbf{Question}: In what year did Massachusetts first require children to be educated in schools?\\
& &\textbf{Answer}: 1852\\
\cmidrule{2-3}
       & \multirow{3}{*}{QA-SRL}    &\textbf{Context}:the race is in mixed eights , and usually held in late february / early march.\\
    &  &\textbf{Question}:when is something held ? \\
    &   &\textbf{Answer}:in late february / early march \\
    \cmidrule{2-3}
     & \multirow{3}{*}{QA-ZRE}     &\textbf{Context}:travis hamonic ( born august 16 , 1990 ) is a canadian professional ice hockey...\\
 &  &\textbf{Question}:what team does travis hamonic belong to ? \\
  &  &\textbf{Answer}:new york islanders\\
\midrule
\multirow{9}{*}{Sequence Generation}    & \multirow{3}{*}{CNN/DM}   &\textbf{Context}:( cnn ) governments around the world are using the threat of terrorism... \\
& &\textbf{Question}:what is the summary ?\\
& &\textbf{Answer}:amnesty ' s annual death penalty report catalogs encouraging signs... \\
\cmidrule{2-3}
 & \multirow{3}{*}{WOZ} &\textbf{Context}:what is the phone number and postcode of a cheap restaurant in the east part of town ?...\\
 & &\textbf{Question}:what is the change in state ?\\
& &\textbf{Answer}:price range : cheap , area : east ; phone , postcode \\
\cmidrule{2-3}
 & \multirow{3}{*}{WikiSQL} &\textbf{Context}:the table has columns player , no . , nationality , position , years in toronto...\\
 & &\textbf{Question}:what is the translation from english to sql ?  \\
& &\textbf{Answer}:select nationality from table where player = terrence ross \\
\midrule
\multirow{8}{*}{Text Classification}  & \multirow{3}{*}{SST}      & \textbf{Context}:no movement , no yuks , not much of anything .\\
& & \textbf{Question}:is this review negative or positive ? \\
& &\textbf{Answer}: negative\\
\cmidrule{2-3}
 & \multirow{4}{*}{MNLI}     &\textbf{Context}:premise:yeah i i think my favorite restaurant is always been the one closest you... \\
& & \textbf{Question}:hypothesis:i like him for the most part , but would still enjoy seeing someone beat him. \\
& &  - - entailment , neutral , or contradiction ?\\
& &\textbf{Answer}: entailment\\
\bottomrule
\end{tabular}}
\caption{Samples extracted from different decaNLP tasks. Each task contains a context, a question and an answer. Note that SQuAD is in the QA task set as well.}
\label{tab:cases-deca}
\end{table*}

\begin{table*}[!t]
\centering
\small
\scalebox{0.9}{
\begin{tabular}{c|c|l}
\toprule
Format    & Dataset  & Case \\
\midrule
\multirow{9}{*}{Extractive} & \multirow{3}{*}{SQuAD} & \textbf{Context}: (Private\_school) Private schooling in the United States has been... \\
& &\textbf{Question}: In what year did Massachusetts first require children to be educated in schools?\\
& &\textbf{Answer}: 1852\\
\cmidrule{2-3}
       & \multirow{3}{*}{NewsQA}    &\textbf{Context}:ABECHE, Chad (CNN) -- Most of the 103 children that a French charity...  \\
    &  &\textbf{Question}:WHO ARE UNDER ARREST IN CHAD? \\
    &   &\textbf{Answer}:Three French journalists, a seven-member Spanish flight crew and one Belgian \\
    \cmidrule{2-3}
     & \multirow{3}{*}{Quoref}     &\textbf{Context}:(Blast of Silence) Frankie Bono, a mentally disturbed hitman from Cleveland... \\
 &  &\textbf{Question}:What is the first name of the person who follows their target to select...? \\
  &  &\textbf{Answer}:Frankie \\
\midrule
\multirow{9}{*}{Abstractive}    & \multirow{3}{*}{NarQA}   &\textbf{Context}:The play begins with three pages disputing over the black cloak usually worn by the actor... \\
& &\textbf{Question}:WHO NORMALLY DELIVERS THE OPENING PROLOGUE IN THE PLAY? \\
& &\textbf{Answer}:THE ACTOR WEARING THE BLACK CLOAK \\
\cmidrule{2-3}
 & \multirow{3}{*}{NQOpen} &\textbf{Context}:- cartilage - cartilage cartilage is a resilient and smooth elastic tissue , a rubber...\\
 & &\textbf{Question}:where is each type of cartilage located in the body? \\
& &\textbf{Answer}:many other body components \\
\cmidrule{2-3}
 & \multirow{3}{*}{Drop} &\textbf{Context}:Hoping to rebound from their loss to the Patriots, the Raiders stayed at home for a Week...\\
 & &\textbf{Question}:How many field goals did both teams kick in the first half?  \\
& &\textbf{Answer}:2 \\
\midrule
\multirow{21}{*}{Multiple-Choice}  & \multirow{5}{*}{RACE}      & \textbf{Context}:It's cool, and it's hot, and everyone is doing it. People talk about it often, and friends...\\
& & \textbf{Question}:A blogger is a person \_ . \\
& & (A) who teaches kids bad words (B) who posts songs from the latest bands\\
& &(C) who got drunk last weekend (D) who writes diaries online\\
& &\textbf{Answer}: who writes diaries online\\
\cmidrule{2-3}
 & \multirow{4}{*}{OBQA}     &\textbf{Context}:Null\\
 & &\textbf{Question}:Frilled sharks and angler fish live far beneath the surface of the ocean, which is why they are\\
 & &known as (A) Deep sea animals (B) fish (C) Long Sea Fish (D) Far Sea Animals	Deep sea animals  \\
& &Answer:Deep sea animals \\
\cmidrule{2-3}
   & \multirow{4}{*}{MCTest}     &\textbf{Context}:It was Jessie Bear's birthday. She was having a party... \\
  & &\textbf{Question}:Who was having a birthday? \\
   & &(A) Jessie Bear (b) no one (C) Lion (D) Tiger \\
& &\textbf{Answer}:Jessie Bear\\
\cmidrule{2-3}
   & \multirow{4}{*}{SIQA}     &\textbf{Context}:Tracy didn't go home that evening and resisted Riley's attacks\\
     & &\textbf{Question}:What does Tracy need to do before this? \\
   & &(A) make a new plan (B) Go home and see Riley (C) Find somewhere to go \\
& &\textbf{Answer}:Find somewhere to go \\
\cmidrule{2-3}
    & \multirow{4}{*}{Dream}    &\textbf{Context}:M: How long have you been teaching in this middle school? W: For ten years... \\
       & &\textbf{Question}:What's the woman probably going to do? \\
    & &(A) To teach a different textbook. (B) To change her job. (C) To learn a different textbook. \\
& &\textbf{Answer}:To change her job.  \\
\bottomrule
\end{tabular}}
\caption{Samples extracted from different QA tasks. Each task contains a context, a question and an answer.}
\label{tab:cases}
\end{table*}

\begin{algorithm*}[tb]
\caption{Training process of Diana}
\label{alg:algorithm}
\textbf{Input}: prompt-enhanced model $g_{\theta}$, datasets $\{(C_j,Q_j,A_j)\}_{j=1}^{n_i}$ for each task $T_i$ ($i$=$1,\cdots,N$), memory buffer $\mathcal{M}$, general prompt $P_g$, format prompts $\{P_f(F_j)\}_{j=1}^F$, task prompts $\{P_t(T_i)\}_{i=1}^N \cup \{\hat{P}_t(F_j)\}_{j=1}^F$, meta prompts $\{P_m^i\}_{i=1}^M$, task prompt keys $\{\bm{k}_t(T_i)\}_{i=1}^N$, meta prompt keys $\{\bm{k}_m^i\}_{i=1}^M$
\begin{algorithmic}[1] 
\STATE \textbf{Initialize}: $\mathcal{M} \leftarrow \emptyset$ 
\FOR {Each task $T_i$, $i$ = $1,\cdots,N$}
\IF {$\mathcal{M} \neq \emptyset$}
\STATE Calculate cluster centroids $\bm{c}_1, \cdots, \bm{c}_B$ of $\mathcal{M}$
\ENDIF
\FOR {number of training epochs}
\FOR {Each mini-batch $I \in$ $\{(C_j,Q_j,A_j)\}_{j=1}^{n_i} \cup \mathcal{M}$}
\STATE Obtain $\epsilon_k$ by Eq.~\ref{eq:scheduled_sampling}
\FOR {($C, Q, A$) $\in$ $I$}
\STATE Obtain format $F_j$ of ($C, Q, A$)
\STATE Sample $\epsilon, \zeta$ from $U(0,1)$
\IF {$\zeta<\omega$} 
\STATE $P_t(C,Q) \leftarrow \hat{P}_t(F_j)$ \COMMENT{Use task prompt $\hat{P}_t(F_j)$ for unseen tasks}
\ELSIF {$\epsilon<\epsilon_k$}
\STATE $P_t(C,Q) \leftarrow P_t(T_i)$ \COMMENT{Use the golden truth task identity to select task prompt}
\ELSE 
\STATE $P_t(C,Q) \leftarrow P_t(\mathop{{\rm argmin}}\limits_{T_{\tau} \in \{T_1,\cdots,T_i\}}(||\bm{q},\bm{k}_t(T_{\tau})||))$ \COMMENT{Use the inferred task identity to select task prompt}
\ENDIF
\STATE $\mathcal{S}(C,Q) \leftarrow$ indexes of $M'$ meta prompt keys that are closest to $\bm{q}$
\STATE $P_m(C,Q) \leftarrow \mathop{\{P_m^j\}}\limits_{j \in \mathcal{S}(C,Q)}$
\STATE $P(C,Q)$$\leftarrow$$[P_g;P_f(F_j);P_t(C,Q);P_m(C,Q)]$
\STATE Calculate per sample loss $\mathcal{L}_{LM}$ on $g_{\theta}$ and $P(C,Q)$ by Eq.~\ref{eq:lm}
\STATE Obtain negative sample $(C_n,Q_n)$ from $\mathcal{M}$ by Eq.~\ref{eq:neg}
\STATE Calculate per sample loss $\mathcal{L}_t$ on $\bm{k}_t(T_i)$ by Eq.~\ref{eq:metric_learning}
\STATE Calculate per sample loss $\mathcal{L}_m$ on $\{\bm{k}_m^{s_j}\} (s_j \in \mathcal{S}(C,Q))$ by Eq.~\ref{eq:meta_loss1}
\IF {($C, Q, A$) $\in \mathcal{M}$}
\STATE Calculate per sample loss $\mathcal{L'}_m$ on $\{\bm{k}_m^{s_j}\} (s_j \in \mathcal{S}(C,Q))$ by Eq.~\ref{eq:meta_loss2}
\ENDIF
\ENDFOR
\STATE Update $g_{\theta}$ and prompts with accumulated $\mathcal{L}_{LM}$
\STATE Update task prompt keys $\{\bm{k}_t(T_i)\}_{i=1}^N$ with accumulated $\mathcal{L}_t$
\STATE Update meta prompt keys $\{\bm{k}_m^i\}_{i=1}^M$ with accumulated $\mathcal{L}_m$ and $\mathcal{L'}_m$
\ENDFOR
\ENDFOR
\STATE Update $\mathcal{M}$ with $\{(C_j,Q_j,A_j)\}_{j=1}^{n_i}$ according to details in Appendix~\ref{app:mem}
\ENDFOR
\end{algorithmic}
\end{algorithm*}

\label{sec:appendix}
\end{document}